\documentclass[conference]{IEEEtran}
\IEEEoverridecommandlockouts

\usepackage{cite}
\usepackage{amsmath,amssymb,amsfonts}
\usepackage{algorithmic}
\usepackage{multirow}

\usepackage{graphicx}
\usepackage{textcomp}
\usepackage{xcolor}
\usepackage{subcaption}
\def\BibTeX{{\rm B\kern-.05em{\sc i\kern-.025em b}\kern-.08em
    T\kern-.1667em\lower.7ex\hbox{E}\kern-.125emX}}
\usepackage{hyperref}
\hypersetup{
    colorlinks=true,
    allcolors=blue
}

\definecolor{ckgreen}{rgb}{0,0.56,0}

\usepackage{array}
\newcommand{\PreserveBackslash}[1]{\let\temp=\\#1\let\\=\temp}
\newcolumntype{C}[1]{>{\PreserveBackslash\centering}p{#1}}
\newcolumntype{R}[1]{>{\PreserveBackslash\raggedleft}p{#1}}
\newcolumntype{L}[1]{>{\PreserveBackslash\raggedright}p{#1}}
\usepackage{booktabs}
\graphicspath{ {images/} }
\newcommand{\etal}{\textit{et al}.}

\begin{document}
\title{Fruit Quality Assessment with Densely Connected Convolutional Neural Network}

\author{
\IEEEauthorblockN{  
    Md. Samin Morshed\IEEEauthorrefmark{1},
    Sabbir Ahmed\IEEEauthorrefmark{2},
    Tasnim Ahmed\IEEEauthorrefmark{3},
    Muhammad Usama Islam\IEEEauthorrefmark{4}, 
    A.B.M. Ashikur Rahman\IEEEauthorrefmark{8}
                }

\IEEEauthorblockA{\IEEEauthorrefmark{1}\IEEEauthorrefmark{2}\IEEEauthorrefmark{3}Department of Computer Science and Engineering, Islamic University of Technology, Gazipur 1704, Bangladesh}

\IEEEauthorblockA{\IEEEauthorrefmark{4} School of Computing and Informatics, University of Louisiana at Lafayette, Lafayette, LA, USA}

\IEEEauthorblockA{\IEEEauthorrefmark{8} Department of ICS, King Fahd University of Petroleum \& Minerals, Dhahran, Saudi Arabia}

\IEEEauthorblockA{
    \{\IEEEauthorrefmark{1}saminmorshed,
        \IEEEauthorrefmark{2}sabbirahmed,
        \IEEEauthorrefmark{3}tasnimahmed\}@iut-dhaka.edu,
\IEEEauthorrefmark{4}usamaislam1@louisiana.edu,
\IEEEauthorrefmark{8}g202204800@kfupm.edu.sa}
}

\IEEEpubid{
\begin{minipage}[t]{\textwidth}\ \\[10pt]
      \small{
      (Copyright $\copyright$ 2022 IEEE. This work has been submitted to the IEEE for possible publication. Copyright may be transferred without notice, after which this version may no longer be accessible.)}
\end{minipage}
}

\maketitle
\pagestyle{plain}

\begin{abstract}
Accurate recognition of food items along with quality assessment is of paramount importance in the agricultural industry. Such automated systems can speed up the wheel of the food processing sector and save tons of manual labor. In this connection, the recent advancement of Deep learning-based architectures has introduced a wide variety of solutions offering remarkable performance in several classification tasks. In this work, we have exploited the concept of Densely Connected Convolutional Neural Networks (DenseNets) for fruit quality assessment. The feature propagation towards the deeper layers has enabled the network to tackle the vanishing gradient problems and ensured the reuse of features to learn meaningful insights. Evaluating on a dataset of 19,526 images containing six fruits having three quality grades for each, the proposed pipeline achieved a remarkable accuracy of 99.67\%. The robustness of the model was further tested for fruit classification and quality assessment tasks where the model produced a similar performance, which makes it suitable for real-life applications.

\end{abstract}


\section{Introduction}

Fruit classification has emerged as an important aspect in the domain of agriculture, machine learning, and image classification in recent years. Fast and accurate fruit classification is a major challenge to consider to increase efficiency in the farming sector~\cite{bhargava2021fruits, usama2021towards}.
% cite-hasantowards, dubey2015application
Deep learning models, which are basically based on artificial neural networks have shown formidable performances in fruit detection and classification tasks~\cite{ukwuoma2022recent}. 
% hameed2018comprehensive, tripathi2020role,
The potential and prospect of Deep Learning (DL) were further explored using Convolutional Neural Network (CNN) based architectures~\cite{kazi2022determining}. Albeit having substantial prospects, the main challenges that fruit identification research faces involve challenges pertaining to the irregularity of form, size, and variability in color~\cite{ukwuoma2022recent}.  

Classification of fresh and damaged fruits was explored through the means of CNN by Kumar \etal~\cite{kumar2022novel}. While the work carried out by Kazi and Panda~\cite{kazi2022determining} also explored the freshness of fruits, their approach differed by employing the power of transfer learning instead of a vanilla CNN. 
Kumar's work~\cite{kumar2022novel} showed an accuracy of 97.14\% while Kazi \etal~\cite{kazi2022determining} reported having 99\% accuracy, which implies DL models can perform better through transfer learning and fine-tuning. 
% Similarly, Valdez~\cite{valdez2020apple} used YOLOv3 architecture to investigate the suitability in detecting apple defects and eventually reported that in comparison to YOLOv3, while YOLOv3 was better at detecting defects, single shot detectors (SSD) provided a better performance. Research on apple detection and sorting through machine vision was also explored in~\cite{sofu2016design}. 

Siddiqi \etal~\cite{siddiqi2019automated} affirmed the findings reported by Valdez \etal~\cite{valdez2020apple} and reiterated finding better results with SSD
% , thus establishing SSD to be a good contender for evaluating fruit quality. 
The authors of~\cite{fu2020fruit} investigated several networks and proposed a CNN-YOLO induced regression network for fruit quality detection on six types of fruits that aligned with the  work carried out using YOLO in~\cite{valdez2020apple,siddiqi2019automated}. Hussain \etal~\cite{hussain2022simple} provided a dataset worth 10,000 fruits images and proposed a Deep CNN achieving an accuracy of 96\%. 
% Simillarly, research work in~\cite{gill2022fruit} investigated and also proposed DCNN, as well as  Type-II Fuzzy, Teacher-learner based optimization (TLBO), Recurrent Neural Network (RNN), and Long Short-Term Memory (LSTM) based applications for fruit quality understanding.
Meshram \etal~\cite{meshram2022fruitnet} curated a dataset containing more than 19,526 images of highly popular fruits in India with three quality labels, namely good, bad, and mixed quality. A framework named MNet was proposed in~\cite{meshram2021mnet} for reducing fruit misclassification, where the authors curated a dataset having 12,000 images for binary classification and experimented with different state-of-the-art CNN architectures. 
% The research contributed to binary classification with an accuracy of 99.92\% achieved using InceptionV3 architecture. 
% Feature extraction was explored in~\cite{karakaya2019comparative} where the authors utilized histograms, gray level co-occurrence matrices, bag of features and CNNs that was eventually used in the proposed computer vision based framework. 
% Detection of rotten fruits and vegetables by the means of feature extraction through CNNs were also explored by Jana and colleagues~\cite{jana2021detection} where the researchers reported a formidable accuracy of 97.74\% while experimenting with 13599 images of rotten and fresh fruits. Freshness was further explored as a classification problem in~\cite{arunachalaeshwaran2022freshness} where the authors explored hog plum and used augmentation for training purpose and performed grid search as well as k-fold cross validation for hyperparameter tuning.

\section{Methodology}
\subsection{Dataset}

\begin{figure*}[htb]
    \centering
    \subfloat[Apple-good]{
    \includegraphics[width=0.15\textwidth, height=2.5cm]{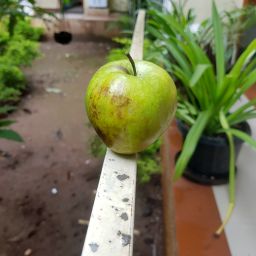}	
    }
    \subfloat[Banana-good]{
    \includegraphics[width=0.15\textwidth, height=2.5cm]{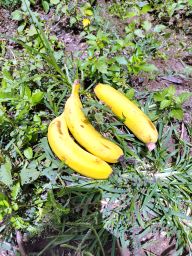}}
    \subfloat[Guava-good]{
    \includegraphics[width=0.15\textwidth, height=2.5cm]{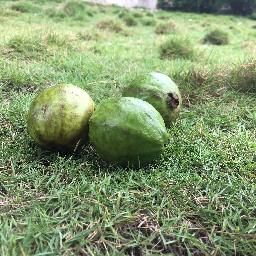}}
    \subfloat[Lime-good]{
    \includegraphics[width=0.15\textwidth, height=2.5cm]{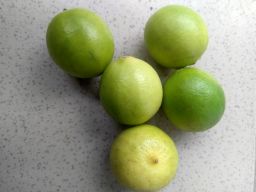} }
    \subfloat[Orange-good]{
    \includegraphics[width=0.15\textwidth, height=2.5cm]{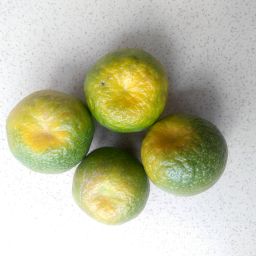}}
    \subfloat[Pomegranate-good]{
    \includegraphics[width=0.15\textwidth, height=2.5cm]{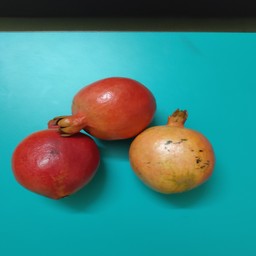}}\\
    
    \subfloat[Apple-bad]{
    \includegraphics[width=0.15\textwidth, height=2.5cm]{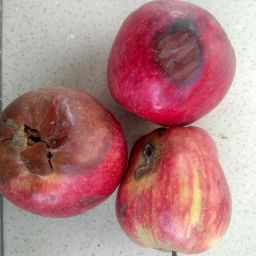}	
    }
    \subfloat[Banana-bad]{
    \includegraphics[width=0.15\textwidth, height=2.5cm]{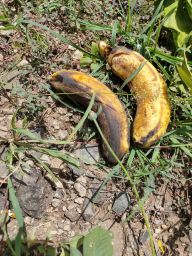}	
    }
    \subfloat[Guava-bad]{
    \includegraphics[width=0.15\textwidth, height=2.5cm]{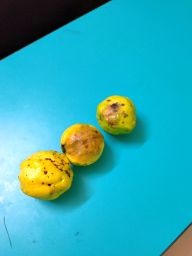}	
    }
    \subfloat[Lime-bad]{
    \includegraphics[width=0.15\textwidth, height=2.5cm]{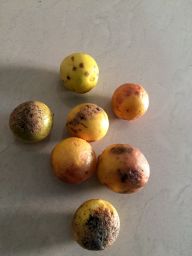}	
    }
    \subfloat[Orange-bad]{
    \includegraphics[width=0.15\textwidth, height=2.5cm]{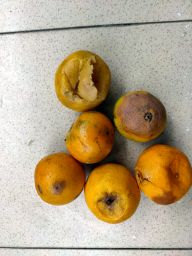}	
    }
    \subfloat[Pomegranate-bad]{
    \includegraphics[width=0.15\textwidth, height=2.5cm]{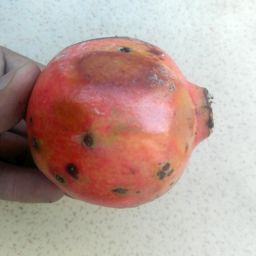}	
    }\\
    
    \subfloat[Apple-mixed]{
    \includegraphics[width=0.15\textwidth, height=2.5cm]{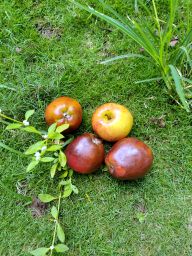}	
    }
    \subfloat[Banana-mixed]{
    \includegraphics[width=0.15\textwidth, height=2.5cm]{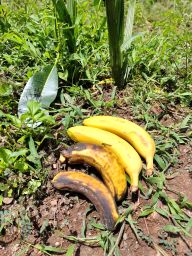}	
    }
    \subfloat[Guava-mixed]{
\includegraphics[width=0.15\textwidth, height=2.5cm]{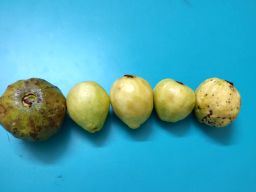}	
    }
    \subfloat[Lime-mixed]{
    \includegraphics[width=0.15\textwidth, height=2.5cm]{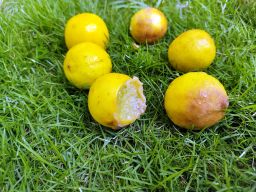}	
    }
    \subfloat[Orange-mixed]{
    \includegraphics[width=0.15\textwidth, height=2.5cm]{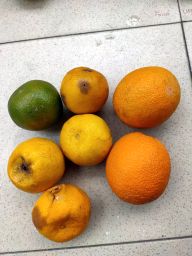}	
    }
    \subfloat[Pomegranate-mixed]{
    \includegraphics[width=0.15\textwidth, height=2.5cm]{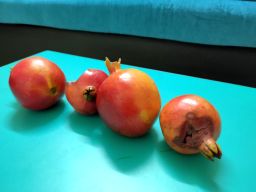}	
    }
    \caption{Sample images of the FruitNet dataset}
    \label{fig:dataset}
\end{figure*}

The experiments were conducted on the `FruitNet' dataset proposed by Meshram \etal \cite{meshram2022fruitnet}, which contains images of fruits along with quality levels. There are six different fruits (apple, banana, guava, lime, orange, and pomegranate) and for each fruit, three types of quality information are available (good, bad, and mixed). The dataset contains 19,526 images belonging to a total of 18 classes. Detailed information about the class distribution is available in Table \ref{tab:dataset_dist}.

The samples were captured with a high-resolution smartphone camera under varying lighting and background conditions. Samples of the same fruit belonging to different quality levels contain a high degree of visually similar features which is also an additional challenge for the model to overcome. The samples of different fruits contain similarities in terms of color, shape, and texture pattern. Moreover, samples within the same class contain a high degree of variability due to the diversity of breeds. These challenges make this dataset a suitable one to train intelligent systems that can be applicable to real-life problems.  
In this work, we utilized this dataset for three different tasks, which are, `Fruit Classification' (6 classes), `Quality Assessment' (3 classes), and `Fruit Classification with Quality Levels' (18 classes). \figureautorefname~\ref{fig:dataset} contains one sample from each of the classes of the dataset. 

\begin{table}[b]
    \centering
    \caption{Distribution of Samples in the Dataset}
    \label{tab:dataset_dist}
    
    \begin{tabular}{lcccc}
        \toprule
        \textbf{Class Label} & \textbf{Good} & \textbf{Bad} & \textbf{Mixed} & \textbf{Total}  \\ 
        \midrule
        Apple & 1149 & 1141 & 113 & 2403 \\ 
        Banana & 1113 & 1087 & 285 & 2485 \\ 
        Guava & 1152 & 1129 & 148 & 2429 \\ 
        Lime & 1094 & 1085 & 278 & 2457 \\ 
        Orange & 1216 & 1159 & 125 & 2500 \\ 
        Pomegranate & 5940 & 1187 & 125 & 7252 \\ 
        \midrule
        Total & 11664 & 6788 & 1074 & 19526 \\ 
        \bottomrule
    \end{tabular}
\end{table}

\subsection{Model Description}
Convolutional Neural Networks (CNNs) are being considered the gold standard to deal with image classification tasks in recent times \cite{ahmed2019ocrBangla, arowa2022voiceCommand}.
Since the introduction of Alexnet, CNNs outperformed all existing algorithms by a significant margin and there has been a great deal of interest in investigating various neural network topologies and experimenting with various strategies such as expanding the width or depth of neural networks \cite{ashikur2022twoDecades}. Deeper networks are thought to perform better in general due to their increased learning capacity. However, with the increasing depth of the architecture, the problem of vanishing gradient arises.

To address these difficulties, alternative architectures with residual connections were developed. Densely Connected Deep Neural Networks (DenseNets) handle the same issue by having the output from each layer concatenated to all subsequent layers \cite{Huang_2017_CVPR_DenseNet}. 
% A traditional feedforward architecture, passes the output from one layer to the next layer, and as the it grows deeper, it will have access to only higher level features rather than those from lower level one that are closer to the input. 
In DenseNets, every subsequent layer receives information from all the preceding layers of the chain. These values are concatenated to ensure that no information is lost. This concatenated feature map is run through a composite function that includes Batch Normalization, Relu, and $3\times3$ Convolution. The result is then passed on to the next layer and the process is repeated.

In a deep network, this concatenation may result in a very large number of parameters. As a result, the design is divided into many dense blocks, each separated by a transition block. Downsampling occurs in the transition block to prevent blowing up. The transition layer consists of batch normalization layers, a $1\times1$ convolution layer, and a $2\times2$ average pooling layer. Based on the number of layers, several variants of the basic DenseNet architectures are available, where DenseNet201 has achieved the highest accuracy in the ImageNet dataset. Hence, it has been chosen as the backbone. 
% model.

\subsection{Data Augmentation}
As described in \tableautorefname~\ref{tab:dataset_dist}, the dataset contains class imbalance since the number of samples for mixed quality is significantly lower compared to the good or bad quality classes. As Deep CNNs require a huge amount of data to automatically find the hidden patterns, data augmentation can be extremely useful in this regard to ensure that the model has the opportunity to learn to its fullest capacity. However, traditional augmentation techniques follow policies like undersampling each class to the number of samples in the least populated class or oversampling each class to expand to a threshold. However, 
%both approaches have their drawbacks, as 
undersampling has the risk of losing critical samples, and oversampling can extend the samples only to some extent. Hence, instead of increasing/decreasing the samples before the training phase, we adopted runtime augmentation, which augments the samples with random strength in each epoch, ensuring providing new challenges to the model \cite{ahmed2022lessIsMore}. 

To reflect the real-world scenarios, we have used augmentations like random rotation, horizontal \& vertical flipping, height \& width shifting, and shearing with a random strength. Multiple of these augmentations are applied to the samples allowing a new version to be presented in front of the model each time it is loaded during training. 
% This removes the chances of overfitting as it has no possibility to visit the same samples in every epoch.

\section{Results and Discussions}

\subsection{Experimental Setup}
The proposed pipeline was trained in a Google Colaboratory environment on a NVIDIA Tesla T4 GPU having a virtual memory of 15GB. The input images were resized to a dimension of $256 \times 256 \times 3$. The split ratio of 60:20:20 was followed corresponding to the training, validation, and testing set. The model was trained with a batch size of $32$ to satisfy the available GPU memory. Training was allowed up to $1000$ epochs with early stopping to restrict if no significant improvement was found within 5 epochs. For all of the tasks, the Adam optimizer was used and the Learning rate was set to 0.0001 along with a decay value of 0.1 after 5 patient epochs. 
% We have chosen accuracy, precision, recall, and F-1 score as the evaluation criteria for our proposed architecture.

% \subsection{Evaluation Metrics}
% In our experiments, we have chosen accuracy, precision, recall, and F-1 score as the evaluation criteria for our proposed architecture. Accuracy is a widely used performance metric, but can be a  bit misleading at times, when referring to an imbalanced dataset. Experiments using an imbalance dataset might achieve very high accuracy despite having low accuracy for the smaller classes. To avoid this issue, we have considered precision and recall. Because precision and recall are about exactness and completeness, respectively, they are biased as well. For this reason, we have considered the F-1 score as another performance metric to evaluate our model. 

\subsection{Performance of Different Baseline Architectures}

To choose the best-suited model, a number of state-of-the-art CNN models were evaluated, which were pretrained for the large-scale image classification task and fine-tuned using the FuritNet dataset. With the utilization of transfer learning, the models were already capable of learning complex patterns, ensuring faster convergence. With the goal of picking the best-suited model, we only modified the softmax layer.
% and trained without introducing any augmentation. 
% The architectures were picked considering their commendable performance in large-scale image classification tasks. 
Due to the high learning capacity of these models, all combinations achieved an accuracy above 97\%. Surprisingly, MobileNetV2 produced an accuracy of 98.62\% which is higher than deeper architectures such as ResNet152 and InceptionResNetV2. However, only DenseNet201 architecture achieved an accuracy of more than 99\%. Furthermore, the number of training parameters for DenseNet201 does not impose high computational resources, making it the most suitable model for fruit quality assessment. 

We further enhanced the capability of the best model with Data augmentation techniques. Our idea of runtime augmentation improved the accuracy of the DenseNet201 architecture up to 99.67\% for fine-grain fruit quality assessment. To ensure robustness, two new tasks were introduced; namely, fruit classification (6 class) and quality classification (3 class). In the first task, the model had to predict the class of fruit, and in the second one, it had to predict whether the fruit is of good, bad, or mixed quality. The model, respectively, achieved 99.7\% and 99.67\% accuracy for the two additional tasks.

\begin{table}[tb]
    \centering
    \caption{Performance Comparison of Baseline Architectures}
    \label{tab:performance_comparison}
    \begin{tabular}{L{2.5cm} C{1.5cm} C{3cm}}
    \toprule
    \textbf{Architecture} & \textbf{Accuracy (\%)} & \textbf{Trainable Parameters (millions)}\\
    \midrule

    ResNet152 & 97.86 & 58.26\\
    InceptionResNetV2 & 97.91 & 54.3\\
    EfficientNetV2B0 & 97.95 & 5.88\\
    VGG16 & 98.6 & 14.72\\
    MobileNetV2 & 98.62 & 2.25\\
    InceptionV3 & 98.8 & 21.81\\
    Xception & 98.98 & 20.84\\
    DenseNet201 & \textbf{99.26} & 18.13\\
    \midrule
    % DenseNet201 + aug & 99.67\\
    \end{tabular}
    
\end{table}

\subsection{Class-wise Performance \& Error Analysis}

\begin{figure*}[htb]
    \centering
    \includegraphics[width=\textwidth]{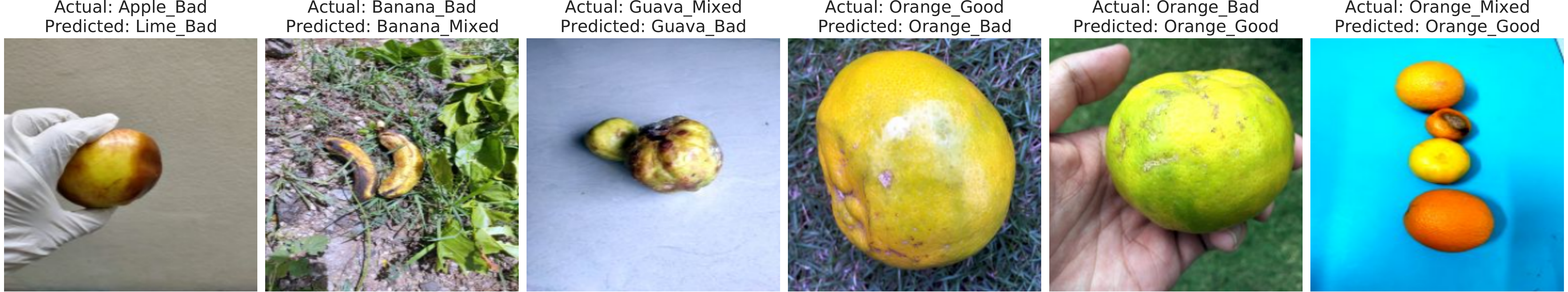}
    \caption{Misclassified samples}
    \label{fig:misclassified}
\end{figure*}

Table \ref{tab:classwise_performance} illustrates the class-wise performance  along with the number of samples of individual classes in the test set. Although accuracy is a very well-known performance metric, it is also known for its sensitivity toward class imbalance. For a more robust analysis, we have considered precision, recall, and F-1 score for each class. Furthermore, to get an insight into the class-separability of our proposed model, the area under the receiver operating characteristics curve (AUC-ROC) is also investigated. The AUC-ROC score can be used to evaluate a classifier's performance by describing the trade-off between the True Positive Rate (TPR) and the False Positive Rate (FPR) using several probability thresholds.
% A perfect classifier will have a ROC with 100\% true positives and no false positives. In general, we count the number of positive classifications as the rate of false positives increases.
Our model achieved an AUC-ROC score of 1 for the majority of the classes, as indicated in Table \ref{tab:classwise_performance}. Scores for the remaining classes are fairly high, showing that class separability is satisfactory. Figure \ref{fig:misclassified} depicts a few of the misclassified samples, and it can be observed that the majority of them can also perplex human decisions regarding the predicted class.

\begin{table}[tb]
\centering
\caption{Performance of Individual Class}
\label{tab:classwise_performance}
\resizebox{\columnwidth}{!}{%
\begin{tabular}{llccccc}
\toprule
\textbf{Fruit}               & \textbf{Quality} & \textbf{Precision} & \textbf{Recall} & \textbf{F1-Score} & \textbf{AUC-ROC} & \textbf{Support} \\ \midrule
\multirow{3}{*}{Apple}  & Bad   & 1.0000 & 0.9956 & 0.9978 & 1.0000 & 229  \\ 
% \cline{2-7} 
                        & Good  & 1.0000 & 1.0000 & 1.0000 & 1.0000 & 231  \\ 
                        % \cline{2-7} 
                        & Mixed & 1.0000 & 1.0000 & 1.0000 & 1.0000 & 24   \\ \midrule
\multirow{3}{*}{Banana} & Bad   & 0.9909 & 0.9954 & 0.9931 & 0.9973 & 218  \\ 
% \cline{2-7} 
                        & Good  & 1.0000 & 1.0000 & 1.0000 & 1.0000 & 224  \\ 
                        % \cline{2-7} 
                        & Mixed & 0.9821 & 0.9649 & 0.9735 & 0.9998 & 57   \\ \midrule
\multirow{3}{*}{Guava}  & Bad   & 0.9913 & 1.0000 & 0.9956 & 0.9999 & 227  \\ 
% \cline{2-7} 
                        & Good  & 0.9957 & 1.0000 & 0.9978 & 1.0000 & 231  \\ 
                        % \cline{2-7} 
                        & Mixed & 1.0000 & 0.9355 & 0.9667 & 0.9999 & 31   \\
                        \midrule
\multirow{3}{*}{Lime}   & Bad   & 0.9954 & 1.0000 & 0.9977 & 0.9999 & 217  \\ 
% \cline{2-7} 
                        & Good  & 1.0000 & 1.0000 & 1.0000 & 1.0000 & 220  \\ 
                        % \cline{2-7} 
                        & Mixed & 1.0000 & 1.0000 & 1.0000 & 1.0000 & 57   \\ \midrule
\multirow{3}{*}{Orange} & Bad   & 0.9830 & 0.9914 & 0.9872 & 0.9999 & 233  \\ 
% \cline{2-7} 
                        & Good  & 0.9917 & 0.9836 & 0.9877 & 0.9999 & 244  \\ 
                        % \cline{2-7} 
                        & Mixed & 1.0000 & 0.9600 & 0.9796 & 1.0000 & 25   \\ \midrule
\multirow{3}{*}{Pomegranate} & Bad              & 1.0000             & 1.0000          & 1.0000            & 1.0000           & 238              \\ 
% \cline{2-7} 
                        & Good  & 1.0000 & 1.0000 & 1.0000 & 1.0000 & 1188 \\ 
                        % \cline{2-7} 
                        & Mixed & 1.0000 & 1.0000 & 1.0000 & 1.0000 & 25   \\ 
                        \bottomrule
\end{tabular}%
}
\end{table}

% \begin{figure*}[tb]
%     \centering
%     \includegraphics[width=\textwidth]{images/conf_mat}
%     \caption{Confusion Matrix}
%     \label{fig:conf_mat}
% \end{figure*}

\begin{figure*}[htb]
\centering
\begin{subfigure}{0.5\textwidth}
  \centering
  	\includegraphics[width=0.95\linewidth]{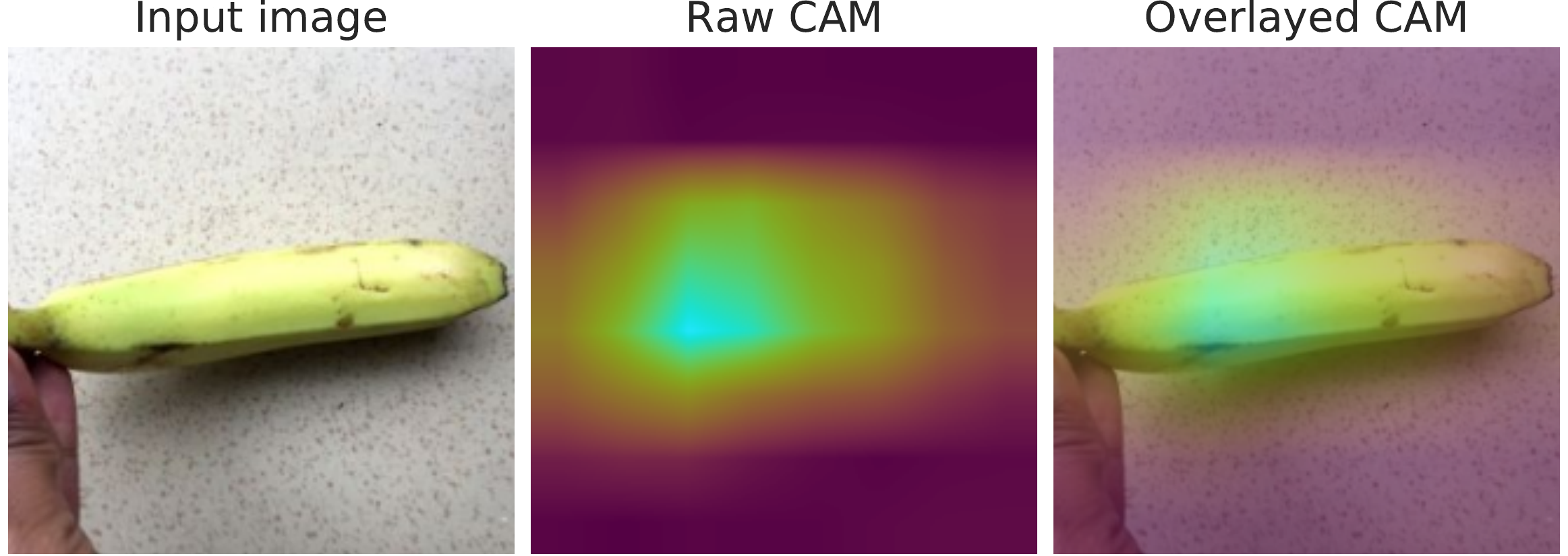}
  \caption{Grad-CAM output for correctly classified `Banana' sample}
  \label{fig:cam_banana}
\end{subfigure}%
\begin{subfigure}{0.5\textwidth}
  \centering
  \includegraphics[width=0.95\linewidth]{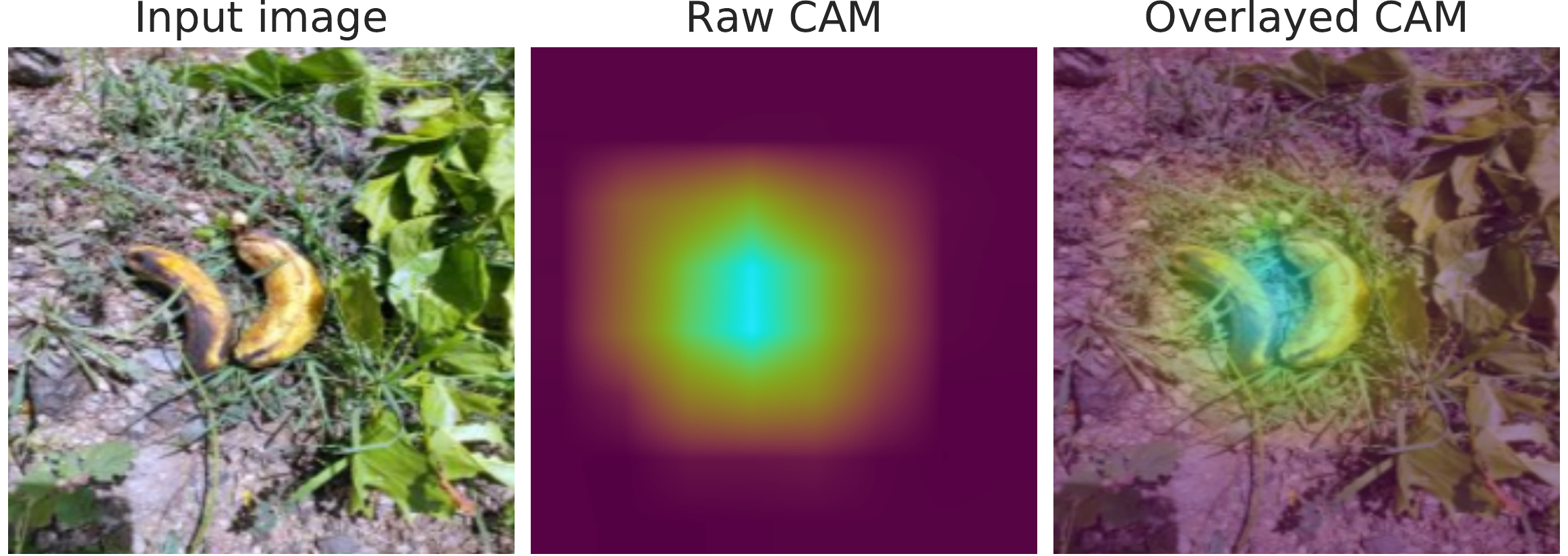}
  \caption{Grad-CAM output for misclassified `Banana' sample}
  \label{fig:cam_misclassified_banana}
\end{subfigure}
\caption{Grad-CAM analysis of the test samples}
\label{fig:gradcam}
\end{figure*}

\subsection{Explainability}
% Explainable artificial intelligence (XAI) is a set of techniques that enable humans to comprehend and trust the results and output of machine learning algorithms. Explainable AI refers to a model's projected influence and probable biases. XAI contributes to the definition of model correctness, fairness, transparency, and results in AI-powered decision-making. 
% Explainability is critical for an organization to create trust and confidence when deploying AI models. 
% Deep neural networks are well known for their complex architecture, which functions in traditional classification or regression applications like a black box architecture. 
An explainability study was conducted to gain insights into the decision-making strategy of the proposed framework. 
Using Grad-CAM \cite{selvaraju2017grad}, we examined the Class Activation Map (CAM) during a forward pass provided by the last convolutional layer. 
%of our proposed architecture. 
The weight for each of the input feature maps is determined by the gradient of the loss with respect to the final convolutional layer. The weighted sum of the activation maps is then up-sampled to the input resolution to build a heatmap that resembles the region that contributes the most to the predicted class. Figure \ref{fig:gradcam} depicts the class activation map of a correctly classified and a misclassified sample, and it is clear that our model provides the highest activation score for the fruit region while disregarding the background.

\section{Conclusion}

Automated quality assessment systems can play an important role in the processing of perishable raw materials by minimizing manual labor. In this work, we proposed a pipeline utilizing the densely connected convolutional neural networks for assessing fruit quality from images. The internal dense connections 
%of the architecture 
alleviated vanishing-gradient problems, strengthened the propagation, and reused features in deeper layers.
%, and reduced the number of parameters to a great extent. 
The capability of the model was enhanced even more by employing run-time augmentation. The robustness of the pipeline was thoroughly investigated for three different tasks- fruit classification, quality assessment, and fine-grained fruit-quality assessment; and it provided a remarkable accuracy of around 99.67\% for all tasks. Furthermore, we provided a detailed class-wise analysis along with an explanation regarding the parts of the weights that excite the decisions.
%for classifying certain classes. 
In the future, this work can be extended by applying image segmentation and object detection models to accurately detect the regions of bad-quality fruits. Assessing the degree of decomposition can be another interesting task to consider.
% \section*{Acknowledgment}

\bibliographystyle{IEEEtran}
\bibliography{References}

\end{document}